\documentclass{article}
\usepackage{spconf,amsmath,graphicx,booktabs,url}
\usepackage{float}
\usepackage{amssymb}
\usepackage{enumitem}
\usepackage{orcidlink}
\hypersetup{
    hidelinks
}

\newif\ifblind

\title{Hydro-Dataset: A Multi-Modal Open-Source Pipeline for Nordic Hydrological Research and Foundation Model Benchmarking}

\ifblind
  \name{Anonymous Author(s)}
  \address{Affiliation withheld for review}
\else

\name{
Getnet Demil\textsuperscript{~1~*}\,\orcidlink{0009-0006-5158-0895}\thanks{Corresponding author. Email: \texttt{getnet.demil@oulu.fi}},
Md Moinul Islam \textsuperscript{1} \,\orcidlink{0000-0003-4185-382X},
Mourad Oussalah \textsuperscript{1}\,\orcidlink{0000-0002-4422-8723}
}

\address{\textsuperscript{1}
Center for Machine Vision and Signal Analysis, University of Oulu, Finland
}

\fi

\begin{document}
%
\maketitle
\begin{abstract}
Despite world-class snow monitoring by the Swedish Meteorological and Hydrological Institute (SMHI) and  Finnish Meteorological Institute (FMI), no station-anchored, multi-parameter snow benchmark dataset exists for the Nordic boreal domain. We present \textit{Hydro-Dataset v1.1}, a reproducible open-source pipeline that harmonizes 582 SMHI meteorological observation stations across nine hydrological winters (2015/16-2023/24) into a CF-1.9 NetCDF product with \textbf{1.8~million annotated station-days}. Six derived snow parameters, such as Snow Water Equivalent (SWE), snow phenology, melt rates, degree-day melt factor, rain-on-snow flags, and freeze-thaw cycles, supply benchmark-ready labels for hydrological foundation model training. ERA5-Land cross-validation (Pearson $R{=}0.78$, bias~$=-262$~mm) confirms observation-anchored SWE provides a more reliable benchmark than reanalysis for machine learning training.
\end{abstract}

\section{Introduction}
\label{sec:intro}

The sub-arctic Nordic region is undergoing rapid climate-driven transformation, with snowpack dynamics governing over 70\% of the annual water cycle~\cite{barnett2005potential,pomeroy2007cold}. Reliable Snow Water Equivalent (SWE) estimates are central to hydropower scheduling, flood risk management, and boreal ecosystem modeling. Yet acquiring trustworthy SWE over large Nordic domains remains an open problem~\cite{luomaranta2019snow}. Two dominant existing approaches each carry fundamental limitations for machine learning applications. Satellite and reanalysis products, including ERA5-Land~\cite{munoz2021era5land} and Globsnow~\cite{takala2011estimating}, provide broad spatial coverage but are known to systematically overestimate SWE in Scandinavian boreal terrain~\cite{mudryk2015characterization}. Cross-validation shows ERA5-Land overestimates by 262~mm relative to station-anchored SWE. 

Recent reviews document the field's shift toward AI and machine learning for SWE retrieval~\cite{schilling2024snow,demil2025advances,nearing2021role}. Concretely, cloud-aware binary snow and cloud segmentation in Sentinel-2 imagery~\cite{demil2025seeing} and multi-sensor CNN-transformer fusion for high-resolution fractional snow cover retrieval~\cite{demil2026xfuse} both demonstrate that pixel-level accuracy gains remain bounded by the quality of ground-truth labels used for training and validation. Yet such labels cannot be reliably sourced from reanalysis products. Point-measurement studies, such as the Alpine density regression~\cite{jonas2009estimating} or machine learning (ML) approaches trained on snow courses~\cite{kratzert2019towards}, deliver accuracy in their target regions but are not directly transferable to the Nordic boreal domain, where taiga-class snowpacks exhibit distinct density dynamics~\cite{sturm2010estimating}. Yet despite dense station networks operated by the Swedish Meteorological and Hydrological Institute (SMHI) and the Finnish Meteorological Institute (FMI), no station-anchored, ML-ready, multi-parameter SWE dataset exists for the Nordic domain. Existing products are either too coarse, regionally biased, or restricted to a single variable.

To close this gap, we release \textit{Hydro-Dataset v1.1}: a reproducible, open-source pipeline producing \textbf{1.8~million annotated station-days} from 582~SMHI MetObs stations across nine hydrological winters (2015/16-2023/24), packaged as a CF-1.9 NetCDF product with co-registered snow process labels ready for multi-task ML training. Our contribution lies \textbf{not} merely in the derivation of the equations themselves, but in providing three properties that existing Nordic datasets don't offer in combination~\cite{takala2011estimating, hall2021modis}, while established methods~\cite{sturm2010estimating,hock2003temperature} serve as validated baselines.
\begin{enumerate}[nosep]
  \item Observation-anchored ground truth spanning 582~stations, empirically validated as a more suitable benchmark than reanalysis products for ML training, achieving Pearson $r{=}0.78$ with reduced bias relative to ERA5-Land,
  \item Six co-registered snow parameters, covering SWE, phenological timing, melt and accumulation rates, degree-day melt factor, rain-on-snow flags, and freeze-thaw cycles, stored alongside raw observations in a single self-contained file, enabling multi-task learning without cross-dataset alignment overhead,
  \item A reproducible pipeline that generates the full suite of derived parameters from raw SMHI station data, with a modular architecture designed to be extended to other Nordic countries' monitoring networks.
\end{enumerate}

This dual contribution, comprising the benchmark dataset and the modular pipeline that sustains it, provides the foundational infrastructure for training next-generation hydrological foundation models in sub-arctic environments~\cite{kratzert2019towards,addor2017camels}.

\section{Data Sources}
\label{sec:data_sources}

\textit{Hydro-Dataset v1.1} fuses four observational data streams into ten variables; Table~\ref{tab:sources} characterizes each variable with its units, tier, record count, and key 9-season statistics.

\textbf{SMHI (Sweden):} Daily snow depth, air temperature, and precipitation records for 582 stations (55--70$^{\circ}$N, 10--30$^{\circ}$E) are fetched via the MetObs REST/JSON API;\footnote{\url{https://www.smhi.se/data}} the companion MESAN 2.5~km reanalysis\footnote{\url{https://www.smhi.se/data/temperatur-och-vind/temperatur/historisk-data-i-grib-format/}} (ETRS89-LAEA, EPSG:3035) defines the kriging output grid.

\textbf{FMI (Finland):} Daily station records are accessible via WFS endpoints\footnote{\url{https://en.ilmatieteenlaitos.fi/open-data}} but are not included in v1.1; they are the primary target for v1.2 integration (Section~\ref{sec:impact}).

\textbf{ERA5-Land (Cross-validation):} The snow depth water equivalent variable (\texttt{sde}) is retrieved from the Copernicus Data Store (CDS) API~\cite{hersbach2020era5,munoz2021era5land} exclusively for independent cross-validation of the derived SWE fields.

\begin{table}[h]
\centering
\caption{Dataset variable characterization (582 SMHI stations, 2015/16--2023/24, 1.82~M station-days). Tier~0: raw SMHI observations; Tier~1: derived parameters. Key values and ranges are 9-season means or min--max across qualifying stations.}
\label{tab:sources}
\resizebox{\columnwidth}{!}{%
\begin{tabular}{llclr}
\toprule
Variable & Unit & Tier & Sample & Key value \\
\midrule
Snow depth             & cm                              & 0 & 1.82M stn-days  & --- \\
Air temperature        & $^\circ$C                       & 0 & 1.82M stn-days  & --- \\
Precipitation          & mm\,d$^{-1}$                    & 0 & 1.82M stn-days  & --- \\
Frozen precipitation   & mm\,d$^{-1}$                    & 0 & 1.82M stn-days  & --- \\
\midrule
SWE                    & mm                              & 1 & 1.82M stn-days  & 94.7 mm (mean peak) \\
Melt rate              & mm\,d$^{-1}$                    & 1 & melt-season days & --- \\
Accumulation rate      & mm\,d$^{-1}$                    & 1 & accum.\ days     & --- \\
DDM                    & mm\,$^\circ$C$^{-1}$d$^{-1}$  & 1 & 9--14\% stations & 2.26--3.53 \\
Rain-on-snow           & events\,season$^{-1}$           & 1 & 4{,}959 events   & 398--721 \\
Freeze-thaw cycles     & count\,season$^{-1}$            & 1 & 5{,}238 records  & 18.3--29.2 \\
\bottomrule
\end{tabular}%
}
\end{table}

\section{Methodology}
\label{sec:methodology}

\begin{figure*}[t]
  \centering
  \includegraphics[width=0.85\linewidth]{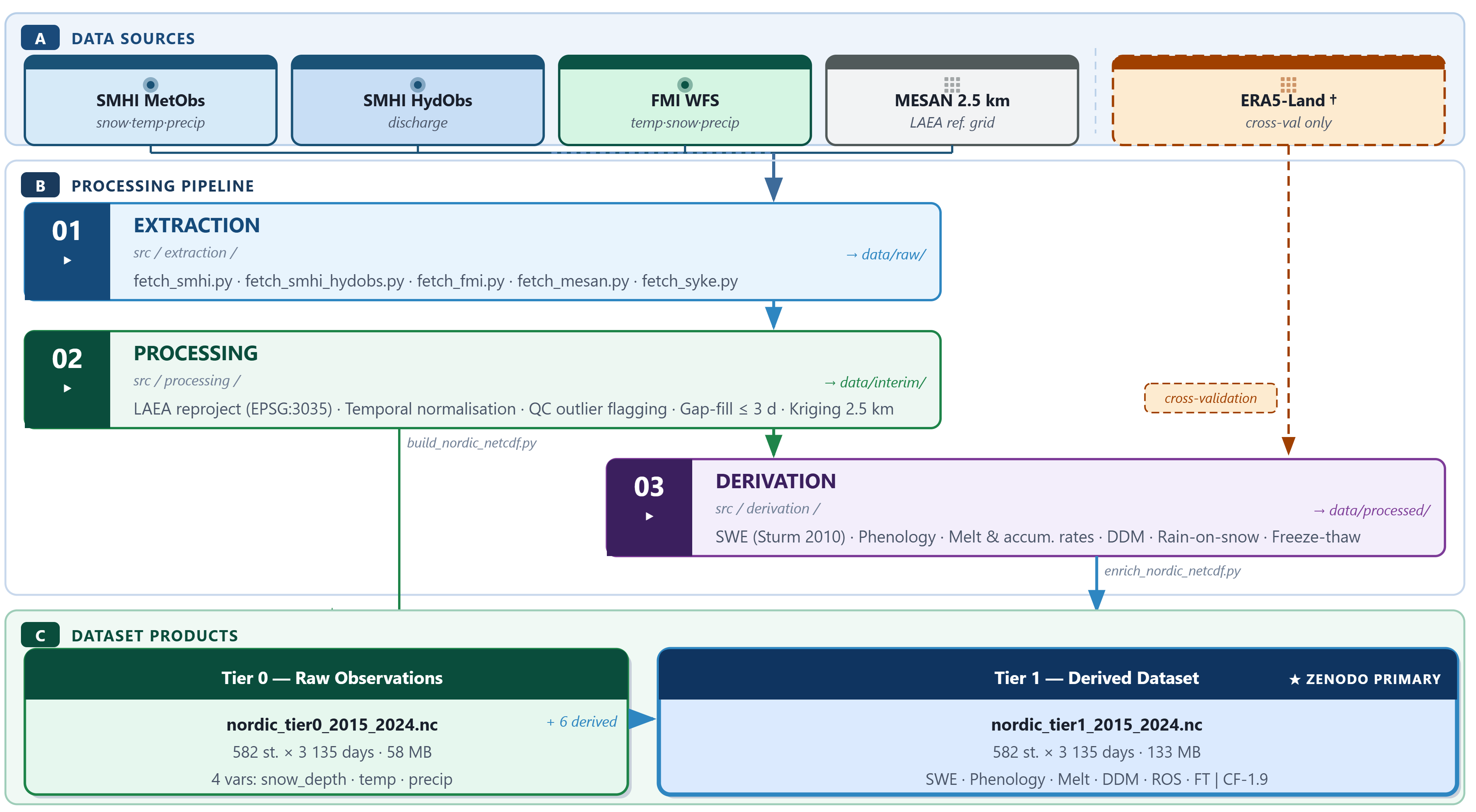}
  \caption{%
    Hydro-Dataset v1.1 pipeline: Extraction (SMHI/FMI APIs) $\to$ Interim storage $\to$ Processing (alignment, QC) $\to$ Derivation (SWE, phenology, melt, ROS, FT, cold content).
  }
  \label{fig:pipeline}
\end{figure*}

The pipeline enforces a strict four-phase sequence (Fig.~\ref{fig:pipeline}): Extraction~$\to$ Interim~$\to$ Processing~$\to$ Derivation, implemented as modular Python scripts. All spatial data are reprojected to ETRS89-LAEA (EPSG:3035) and all temporal data are resampled to daily resolution prior to derivation.

\subsection{Geospatial Alignment and Temporal Resampling}
All station observations are natively referenced in WGS84. Coordinates are reprojected to the 2.5~km LAEA grid used by MESAN via \texttt{pyproj} (\texttt{always\_xy=True}). Flux variables such as precipitation are aggregated to daily totals; state variables such as air temperature are represented as daily means; snow depth retains the end-of-day observation.

\subsection{SWE Derivation}
Among established SWE-from-depth estimation approaches, Yao et al.~\cite{yao2018comparison} demonstrated that snow-density-based models outperform alternatives when daily depth observations are available. We adopt the climate-class density model of Sturm et al.~\cite{sturm2010estimating} in preference to the Swiss Alpine regression of Jonas et al.~\cite{jonas2009estimating}, as the \textit{taiga} class is specifically parameterised for cold continental boreal conditions, the dominant snow regime across the Nordic domain, whereas Jonas et al.\ \cite{jonas2009estimating}, calibrated exclusively for altitude-dependent Alpine snowpacks. The density model is expressed as:
\begin{equation}
\rho(h, d) = (\rho_\mathrm{max} - \rho_0)\bigl[1 - e^{-k_1 h - k_2 d}\bigr] + \rho_0
\label{eq:sturm}
\end{equation}
where $h$ denotes snow depth (cm), $d$ is the day-of-season (days since 1~October), and $\rho_0$, $\rho_\mathrm{max}$, $k_1$, $k_2$ are class-specific empirical constants. For the taiga class, these take values $\rho_0{=}0.217$, $\rho_\mathrm{max}{=}0.490$~g,cm$^{-3}$, $k_1{=}0.0$, and $k_2{=}0.015$~d$^{-1}$. Propagating this density range to the Dalarna pilot peak depth (0.848~m) yields a SWE bound of 184--415~mm; the kriged peak of 185~mm falls near $\rho_0$, consistent with the moderately compacted taiga snowpack at peak accumulation.

\subsection{Spatial Interpolation via Ordinary Kriging}
The primary v1.1 product is a point-station NetCDF file containing all 582 station time series. As a pilot demonstration of the spatial interpolation module, station-point SWE and snow depth are interpolated onto the 2.5~km LAEA grid over the Dalarna region (58$\times$68 cells; 3,944 LAEA cells) using ordinary kriging~\cite{cressie1990origins} with an empirically fitted spherical variogram, implemented via \texttt{pykrige}. Ordinary kriging is preferred over simpler inverse-distance weighting because it yields an unbiased minimum-variance estimate and produces a kriging variance field that explicitly quantifies spatial interpolation uncertainty~\cite{magnusson2014assimilation}. Time steps with fewer than three valid stations within the Dalarna extent are assigned NaN.

\subsection{Snow Phenology}
Per-station seasonal indices are derived from the snow depth time series and comprise four metrics: onset day-of-season, defined as the first day of at least three consecutive snow-covered days with depth$>$1~cm; melt-out day; snow cover duration; and peak SWE day-of-season. These indices characterize the annual snow regime independently of magnitude, providing phenologically informative targets for training.

\subsection{Melt Dynamics and Degree-Day Factor}
Daily snowmelt and accumulation rates are computed from finite differences of the SWE time series~\cite{hock2003temperature}:
\begin{align}
M_t &= \max\bigl(0,; -(SWE_t - SWE_{t-1})\bigr) \label{eq:melt}\\
A_t &= \max\bigl(0,; SWE_t - SWE_{t-1}\bigr)
\end{align}
The degree-day melt factor (DDM) is computed as the per-station median ratio of melt rate to positive air temperature, following the formulation of Hock~\cite{hock2003temperature}:
\begin{align}
\mathrm{DDM} = \mathrm{median}!\left(\frac{M_t}{T_t}\right) \label{eq:ddm}\\ 
M_t \geq 0.5~\mathrm{mm,day}^{-1}, T_t > 0^\circ\mathrm{C}
\end{align}
A minimum of three qualifying day pairs is required. Stations failing this threshold are assigned NaN.

\subsection{Rain-on-Snow Detection}
A binary rain-on-snow (ROS) event flag is raised when three conditions hold simultaneously~\cite{rennert2009soil}:
\begin{equation}
h_\mathrm{snow} > 0.01~\mathrm{m}\wedge T > 0^\circ\mathrm{C} \wedge P > 0.5~\mathrm{mm,day}^{-1}
\end{equation}
ROS events are recognized drivers of rapid snowmelt, downstream flooding, and ecosystem disruption~\cite{rennert2009soil}, constituting rare-event labels of high value for foundation model training.

\subsection{Freeze–Thaw Cycles and Cold Content}
Freeze-thaw cycles are enumerated as sign changes in the daily air temperature series, with days recording exactly zero degrees inheriting the sign of the preceding day to avoid artificial double-counting. Snowpack cold content quantifies the energy deficit required to bring the snowpack to 0$^\circ$C~\cite{pomeroy2007cold}:
\begin{equation}
CC = 10^{-6},\rho_\mathrm{ice}, c_\mathrm{ice}, h, |\min(T_\mathrm{air},, 0)| \quad [\mathrm{MJ,m^{-2}}]
\end{equation}
adopting $\rho_\mathrm{ice}{=}917$~kg,m$^{-3}$ and $c_\mathrm{ice}{=}2090$~J,kg$^{-1}$,K$^{-1}$.

\subsection{Runoff Potential}
Daily surface runoff potential is estimated using the Antecedent Precipitation Index (API) recession model:
\begin{align}
\mathrm{API}t &= k \cdot \mathrm{API}{t-1} + P_t \\
RC_t = \frac{\mathrm{API}_t}{\mathrm{API}_t + S}
Q_t &= (P_t + M_t) \cdot RC_t
\end{align}
where $M_t$ is the daily snowmelt rate from Eq.~(\ref{eq:melt}), $k{=}0.85$ is the recession constant, and $S{=}50$~mm is the saturation deficit. Both are representative empirical values for boreal forest catchments~\cite{pomeroy2007cold}.

\subsection{Cross-Validation Against ERA5-Land}
ERA5-Land snow depth water equivalent is retrieved via the CDS API and bi-linearly interpolated to station locations using Scipy's \texttt{RegularGridInterpolator}. Skill metrics, comprising RMSE, bias, Pearson $r$, and the Kling-Gupta Efficiency (KGE)~\cite{gupta2009decomposition}, are evaluated over all snow-present station-day pairs, defined as days with $h_\mathrm{snow} > 0$.

\section{Research Impact and Versionability}
\label{sec:impact}

\subsection{Positioning Against Existing Datasets}

Table~\ref{tab:comparison} situates Hydro-Dataset v1.1 against the dominant existing SWE and snow monitoring products. No existing Nordic product combines observation-anchored SWE with co-registered process parameters in an open, reproducible pipeline.

\begin{table}[h]
\centering
\caption{Comparison of Hydro-Dataset v1.1 against existing snow/SWE products.}
\label{tab:comparison}
\resizebox{\columnwidth}{!}{%
\begin{tabular}{lccccc}
\toprule
Property & \textbf{Ours} & ERA5-Land & Globsnow~\cite{takala2011estimating} & MODIS MOD10A1~\cite{hall2021modis} & Jonas~\cite{jonas2009estimating} \\
\midrule
Type              & Station obs.  & Reanalysis     & Sat.+station   & Satellite      & Station reg. \\
Domain            & Nordic        & Global         & N.~Hemisphere  & Global         & Swiss Alps \\
Spatial res.      & 582 points    & 9~km           & 25~km          & 500~m          & Point \\
Temporal res.     & Daily         & Hourly         & Daily          & Daily          & Daily \\
Period            & 2015/16--2023/24 & 1950--present  & 1979--2020     & 2000--present  & 1995--2009 \\
SWE variable      & Yes           & Yes            & Yes            & No             & Yes \\
Derived params    & 6             & 0              & 0              & 0              & 0 \\
SWE bias vs.\ Ours & Ref.          & $-$262~mm      & Overestimates  & N/A            & Not applicable \\
Format            & CF-1.9/Xarray & GRIB2          & NetCDF         & HDF4           & CSV \\
Open pipeline     & \checkmark    & $\times$       & $\times$       & $\times$       & $\times$ \\
\bottomrule
\end{tabular}%
}
\end{table}

\subsection{Foundation Model Readiness}
The dataset is designed as training infrastructure for hydrological foundation models. The field's shift toward AI-driven SWE retrieval~\cite{schilling2024snow,nearing2021role} creates immediate demand for labeled multi-parameter Nordic data that satellite and reanalysis products cannot reliably provide~\cite{mudryk2015characterization}. Large-sample hydrology has demonstrated that model performance scales with the volume and geographic diversity of labeled training data~\cite{kratzert2019towards,addor2017camels}. The Tier~1 product delivers \textbf{1.8~million annotated station-days}, each carrying four harmonized observation variables and six co-registered derived snow parameters at matching stations and time steps, eliminating cross-dataset alignment overhead. No existing public product offers co-registered, observation-anchored labels at this scale for the Nordic domain.

\subsection{Reproducibility and Planned Extensions}
The script-driven architecture ensures the dataset can be fully regenerated from raw API sources with a single command (\texttt{make enrich-netcdf}). Future releases expand the Nordic footprint and adopt the \textit{AquaNord} product family name to reflect pan-Nordic scope. \textbf{AquaNord-SE v1.2} incorporates approximately 150 FMI stations, for which extraction scripts are already implemented, expanding coverage to \textbf{\textasciitilde730 stations} and \textbf{\textasciitilde2.2~million station-days}~\cite{luomaranta2019snow}; \textbf{AquaNord-Nordic v1.3} extends the network to Norwegian (met.no) and Danish (DMI) stations, achieving pan-Nordic coverage; \textbf{AquaNord-Nordic v2.0} integrates H-SAF satellite snow cover and ASCAT soil moisture as additional derivation modules.

\subsection{Limitations}

\textbf{Validation.} Cross-validation is conducted against ERA5-Land, a product we argue is systematically biased. While Pearson $r{=}0.78$ confirms temporal agreement, independent validation against SMHI manual snow course surveys remains outstanding and is planned for v1.2. Runoff potential similarly awaits validation against HydObs discharge records, pending station-to-catchment matching.

\textbf{Model simplifications.} The taiga density class is applied uniformly; southern and coastal maritime snowpacks may be better represented by the maritime class. Cold content uses air temperature as a bulk snow temperature proxy, appropriate for continental taiga but unreliable for isothermal snowpacks during late-season melt.

\textbf{Spatial coverage.} Version~1.1 covers Sweden only (582 SMHI stations); FMI and met.no networks are not yet integrated. DDM estimates are available for only 9-14\% of stations meeting the qualifying-day threshold per season, introducing a southward sampling bias. Full Nordic coverage is the primary objective of v1.2-v1.3.

\section{Results: The Hydro-Dataset Product}
\label{sec:results}

\subsection{Pilot Cohort: Dalarna 2023-24}

As a proof of concept, the full pipeline was executed over the Dalarna catchment (60--61.5$^\circ$N, 13--16$^\circ$E), producing a CF-1.9-compliant NetCDF product spanning 22 SMHI MetObs stations across 213 daily time steps (1 October 2023 - 30 April 2024). The station-indexed dataset contains harmonized snow depth, air temperature, and precipitation, with complete provenance metadata embedded in each variable's \texttt{attrs}.

\subsection{Gridded SWE Interpolation}
Applying the Sturm et al.~\cite{sturm2010estimating} taiga-class model (Eq.~\ref{eq:sturm}) to observed station snow depths yielded per-station SWE estimates for all 213 days. Ordinary Kriging then interpolated snow depth and SWE onto a $58{\times}68$ LAEA grid at 2.5~km resolution (3,944 cells, EPSG:3035). Seasonal peak SWE reached 185~mm and peak snow depth 0.848~m, consistent with Dalarna's interior continental snowpack, which typically exceeds the Nordic-wide mean of 94.7~mm (Table~\ref{tab:benchmark})~\cite{mudryk2015characterization}.

\subsection{ERA5-Land Cross-Validation}
Table~\ref{tab:validation} summarizes the comparison of station-level Sturm SWE against ERA5-Land~\cite{hersbach2020era5,munoz2021era5land} \texttt{sde} bi-linearly interpolated to station locations, evaluated over 2,770 snow-present station-day pairs.

\begin{table}[ht]
\centering
\small
\caption{SWE validation: Sturm (2010, taiga) vs.\ ERA5-Land, Dalarna 2023-24 ($N{=}2{,}770$).}
\label{tab:validation}
\begin{tabular}{lc}
\toprule
Metric & Value \\
\midrule
RMSE & 293 mm \\
Bias (Sturm $-$ ERA5-Land) & $-$262 mm \\
Pearson $r$ & 0.78 \\
KGE & $-$0.13 \\
\bottomrule
\end{tabular}
\end{table}

Pearson $r{=}0.78$ confirms that both products capture the same seasonal dynamics. The negative KGE ($-$0.13) is driven almost entirely by the 262~mm systematic offset; decomposing KGE into its correlation, bias-ratio, and variability-ratio components shows that the penalty originates exclusively from the bias term. Station-anchored and reanalysis SWE thus track the same seasonal cycle but diverge substantially in absolute magnitude, consistent with well-documented ERA5-Land overestimation in Scandinavia~\cite{mudryk2015characterization}. Observation-anchored SWE is therefore more suitable as a benchmark for training and evaluating ML models.

\subsection{Swedish SMHI Station Dataset (v1.1)}
Applying the full pipeline to the SMHI archive produced \textit{nordic\_tier0\_2015\_2024.nc}: a CF-1.9 dataset spanning 582 stations, nine hydrological winters (2015/16 through 2023/24), and 3,135 daily time steps (1 October 2015 - 30 April 2024). Stations with fewer than 30 snow-depth records in any season were excluded, retaining 582 of 594 candidates. The dataset covers the Nordic bounding box (55--70$^\circ$N, 10--30$^\circ$E) and includes four Tier-0 variables: \texttt{snow\_depth}, \texttt{temp\_mean}, \texttt{precip}, and \texttt{frozen\_precip}. The enrichment module (\texttt{enrich\_nordic\_netcdf.py}) augments this with six derived parameters to produce the Tier-1 product (\texttt{nordic\_tier1\_2015\_2024.nc}).

\subsection{Interannual Benchmarking}

Table~\ref{tab:benchmark} reports per-season statistics across all 582 stations. The nine-season mean peak SWE is 94.7~mm; 2017-18 was the heaviest season ($+$36.8~mm anomaly) and 2019-20 the lightest ($-$22.7~mm). The DDM remains stable across seasons (median 2.26-3.53~mm~$^\circ$C$^{-1}$~day$^{-1}$), consistent with the 1-5~mm~$^\circ$C$^{-1}$~day$^{-1}$ range reported for Scandinavian boreal forests~\cite{hock2003temperature}. DDM calibration requires sustained melt-day occurrence ($M_t \geq 0.5$~mm~day$^{-1}$, $T_t > 0^\circ$C for at least three days); only 9-14\% of stations (53-82 of 582) meet this threshold per season, reflecting the physical character of northern taiga winters rather than a data deficiency. ROS events peak in 2023-24 (721 station-days), consistent with anomalously warm and wet conditions across Scandinavia that season.

\begin{table}[h]
\centering
\caption{Per-season SMHI-station benchmark results (582 stations, Sweden, 2015/16--2023/24). PeakSWE = SMHI-station mean peak SWE; Anomaly = departure from 9yr mean; DDM = median degree-day factor across stations meeting the qualifying-day threshold (9--14\% per season); ROS = total rain-on-snow station-days; FT = mean freeze-thaw cycles per station.}
\label{tab:benchmark}
\resizebox{\columnwidth}{!}{%
\begin{tabular}{lcrrrrr}
\toprule
Season & PeakSWE & Anomaly & Duration & DDM & ROS & FT \\
       & (mm)    & (mm)    & (days)   & (mm\,$^\circ$C$^{-1}$d$^{-1}$) & (events) & (cycles) \\
\midrule
2015--16 &  83.7 & $-$12.3 & 108.3 & 2.89 &  678 & 21.5 \\
2016--17 &  72.5 & $-$21.7 & 136.9 & 3.21 &  581 & 29.2 \\
2017--18 & 131.6 & $+$36.8 & 128.9 & 2.95 &  621 & 21.2 \\
2018--19 & 102.0 &  $+$6.8 & 101.7 & 2.42 &  483 & 24.8 \\
2019--20 &  75.7 & $-$22.7 & 118.6 & 2.43 &  398 & 23.0 \\
2020--21 &  98.9 &  $+$6.1 & 104.9 & 2.54 &  479 & 18.3 \\
2021--22 &  87.2 &  $-$6.6 & 129.3 & 2.26 &  461 & 28.8 \\
2022--23 &  92.3 &  $-$1.7 & 135.7 & 2.76 &  541 & 26.1 \\
2023--24 & 108.8 & $+$16.5 & 148.6 & 3.53 &  721 & 22.0 \\
\midrule
\textbf{Mean} & \textbf{94.7} & -- & \textbf{123.7} & \textbf{2.78} & \textbf{551} & \textbf{23.9} \\
\bottomrule
\end{tabular}%
}
\end{table}





\section{Conclusion}
\label{sec:conclusion}
\textit{Hydro-Dataset v1.1} delivers \textbf{1.8~million annotated station-days} across 582 SMHI MetObs stations and nine hydrological winters (2015/16-2023/24), comprising four observation variables and six derived snow parameters in a single CF-1.9 product. ERA5-Land cross-validation (Pearson $r{=}0.78$, bias $=-262$~mm) confirms that station-anchored SWE constitutes more credible ground truth than reanalysis. The nine-season benchmark documents interannual SWE variability (peak range 72.5--131.6~mm) and stable degree-day factors (2.26-3.53~mm~$^\circ$C$^{-1}$~day$^{-1}$), consistent with Scandinavian boreal literature. Full reproducibility is ensured by the script-driven pipeline: \texttt{make enrich-netcdf} regenerates the Tier~1 product from raw SMHI APIs. The 1.8~million co-registered labels position the dataset as foundational training infrastructure for hydrological foundation models~\cite{kratzert2019towards,nearing2021role}. The planned roadmap extends coverage in three stages. AquaNord-SE v1.2 integrates approximately 150 FMI stations, for which extraction scripts are already implemented, expanding to \textbf{$\sim$730 stations} and \textbf{$\sim$2.2~million station-days} across Fennoscandia. AquaNord-Nordic v1.3 adds Norwegian (met.no) and Danish (DMI) networks to achieve full pan-Nordic coverage. AquaNord-Nordic v2.0 targets H-SAF satellite snow products and ASCAT soil moisture for a multimodal extension. Nordic water quality monitoring is addressed in a companion paper.

\ifblind
\else

\section*{Acknowledgment}

This research has been conducted with Flagship Programme funding granted by the Research Council of Finland for Digital Waters Flagship (decision no. 359228) at the University of Oulu.

\fi

\section*{Data Availability}
Hydro-Dataset v1.1 (\texttt{nordic\_tier1\_2015\_2024.nc}), companion benchmark CSVs, and the full pipeline source code are available at
\ifblind
  \url{https://anonymous.4open.science/r/Hydro-Dataset-AE94}%
\else
  \url{}https://github.com/getnetdemil/Hydro-Dataset)%
\fi
 ~(code: MIT licence; data: CC-BY-4.0). The dataset is permanently archived on Zenodo at \\ \ifblind [DOI omitted for review]\else \url{}https://doi.org/10.5281/zenodo.20287300\fi. 
\\ Raw inputs are openly accessible via the SMHI Open Data API (\url{}https://opendata.smhi.se)

\bibliographystyle{IEEEbib}
\bibliography{references}

\end{document}